%% file: aaai2026.tex
\documentclass[letterpaper]{article} 
\usepackage[draft]{aaai2026}  
\usepackage{times}  
\usepackage{helvet}  
\usepackage{courier}  
\usepackage[hyphens]{url}  
\usepackage{graphicx} 
\usepackage{natbib}  
\usepackage{caption} 
\usepackage{algorithm}
\usepackage{algorithmic}

\usepackage{newfloat}
\usepackage{listings}
\usepackage{amsmath}
\usepackage{booktabs}
\usepackage[table]{xcolor}         
\usepackage{colortbl}              
\usepackage{multirow}              
\usepackage{pifont}                
\usepackage{siunitx}

\DeclareCaptionStyle{ruled}{labelfont=normalfont,labelsep=colon,strut=off} 
\floatstyle{ruled}
\newfloat{listing}{tb}{lst}{}
\floatname{listing}{Listing}
\title{HyFedRAG: A Federated Retrieval-Augmented Generation Framework \\for Heterogeneous and Privacy-Sensitive Data}

\author {
    Cheng Qian\textsuperscript{\rm 1},  
    Hainan Zhang\textsuperscript{\rm 2}, 
    Yongxin Tong\textsuperscript{\rm 3}, 
    Hong-Wei Zheng \textsuperscript{\rm 4},
    Zhiming Zheng\textsuperscript{\rm 5}
}
\affiliations {
    \textsuperscript{\rm 1}Beijing Advanced Innovation Center for Future Blockchain and Privacy Computing \\
    \textsuperscript{\rm 2}Institute of Artificial Intelligence, Beihang University, China\\
    \texttt{alpacino@buaa.edu.cn,zhanghainan1990@163.com}
}
\usepackage{bibentry}

\begin{document}

\maketitle

\begin{abstract}
Centralized RAG pipelines struggle with heterogeneous and privacy-sensitive data, especially in distributed healthcare settings where patient data spans SQL, knowledge graphs, and clinical notes. Clinicians face difficulties retrieving rare disease cases due to privacy constraints and the limitations of traditional cloud-based RAG systems in handling diverse formats and edge devices. 
To address this, we introduce HyFedRAG, a unified and efficient Federated RAG framework tailored for Hybrid data modalities. By leveraging an edge-cloud collaborative mechanism, HyFedRAG enables RAG to operate across diverse data sources while preserving data privacy. Our key contributions are: (1) We design an edge-cloud collaborative RAG framework built on Flower, which supports querying structured SQL data, semi-structured knowledge graphs, and unstructured documents. The edge-side LLMs convert diverse data into standardized privacy-preserving representations, and the server-side LLMs integrates them for global reasoning and generation. (2) We integrate lightweight local retrievers with privacy-aware LLMs and provide three anonymization tools that enable each client to produce semantically rich, de-identified summaries for global inference across devices. (3) To optimize response latency and reduce redundant computation, we design a three-tier caching strategy consisting of local cache, intermediate representation cache, and cloud inference cache. Experimental results on PMC-Patients demonstrate that HyFedRAG outperforms existing baselines in terms of retrieval quality, generation consistency, and system efficiency. Our framework offers a scalable and privacy-compliant solution for RAG over structural-heterogeneous data, unlocking the potential of LLMs in sensitive and diverse data environments.

\end{abstract}

\input{1introduction}
\input{2relatedwork}

\input{3model}
\input{4experiments}

\input{5evaluation_Results}

\input{6conclusion}
\bibliography{aaai2026}

\section*{Acknowledgments}
This work was funded by the National Natural Science Foundation of China (NSFC) under Grants No. 62406013, the Beijing Advanced Innovation Center Funds for Future Blockchain and Privacy Computing and the Fundamental Research Funds for the Central Universities. 

\clearpage

\end{document}

%% file: 1introduction.tex
\section{Introduction}

\begin{figure}[!ht]
  \centering
  \includegraphics[width=8cm]{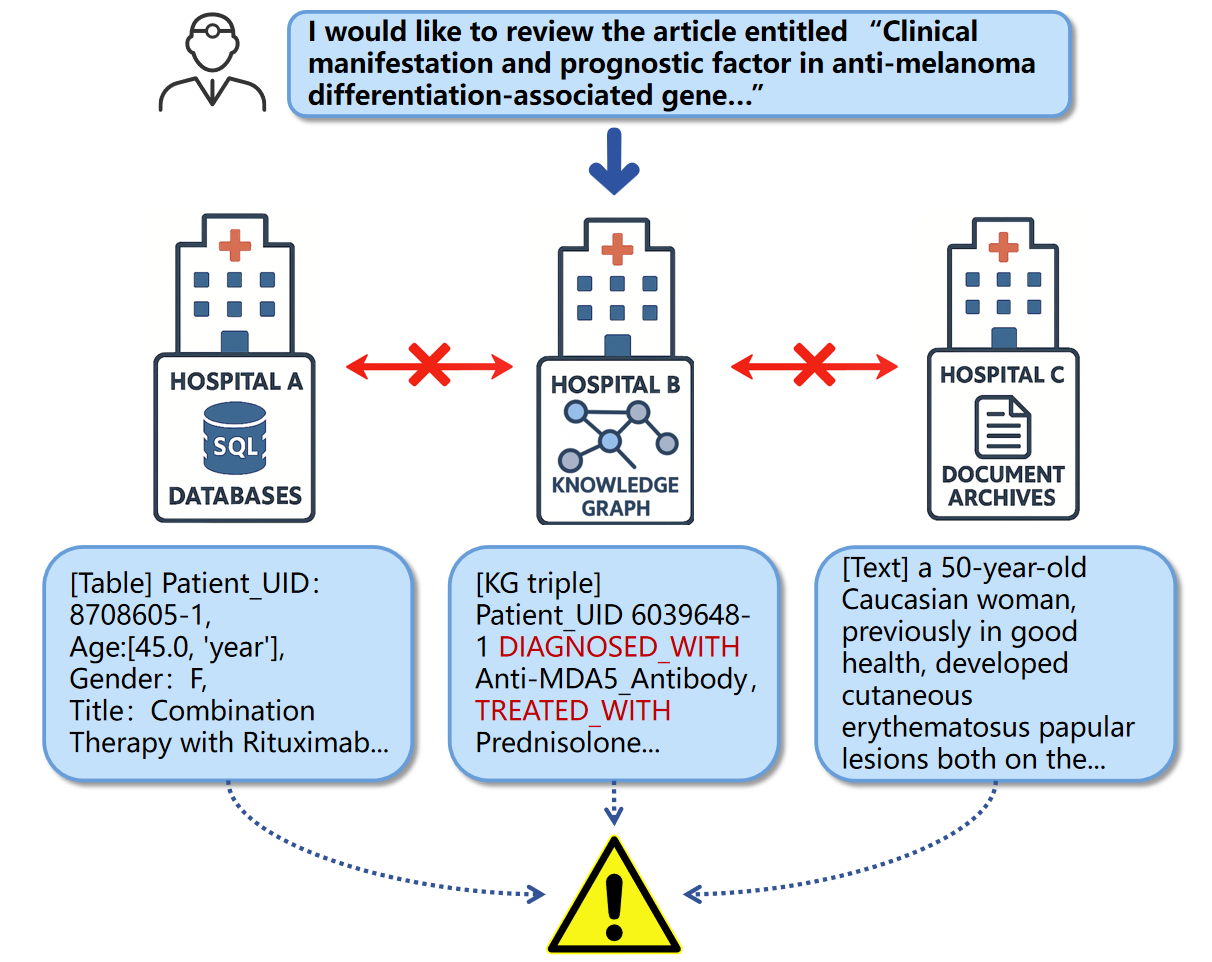}
  \caption{In privacy-sensitive medical environments, the heterogeneous storage formats employed by different hospitals, combined with stringent privacy protection requirements, hinder data interoperability and pose significant challenges for cross‑institutional queries.
}
  \label{fig:intro}
\end{figure}

Large language models (LLMs) have demonstrated remarkable proficiency in natural language understanding and generation\cite{achiam2023gpt}, but they are prone to “hallucinations” caused by missing or outdated knowledge\cite{schick2023toolformer}. Retrieval‑Augmented Generation (RAG)\cite{seo2024replug,zhang2024adacomp} addresses this by invoking external search\cite{zhao2024dense,wang2025maferw} or database queries at inference time, injecting up‑to‑date, domain‑specific information to improve handling of long‑tail entities and specialized terminology. 

However, existing RAG systems typically assume a homogeneous repository (e.g., only text or only graphs), struggling to handle multiple heterogeneous data formats within a single workflow\cite{behnamghader2024llm2vec}, such as structured tables\cite{kong2024opentab}, semi‑structured knowledge‑graph triples\cite{huang2023question}, and unstructured free text. For example, a retriever specialized in table-based retrieval\cite{herzig2021open} cannot be easily applied to knowledge graph-based question answering (QA) tasks\cite{behnamghader2024llm2vec}.

This shortfall is especially problematic in privacy‑sensitive domains like healthcare, where patient data spans electronic health record tables, pathology reports\cite{zhao2022pmc,zheng2024safely}, imaging metadata, and genomic profiles. Each stored under different conventions and protected by regulations such as GDPR and HIPAA that forbid centralizing raw records. For example in Figure~\ref{fig:intro}, clinicians face challenges retrieving rare disease cases due to privacy restrictions on raw data access and the limitations of traditional cloud-based RAG systems in handling diverse edge devices and inconsistent formats. 

To facilitate realistic cross-institution federated RAG, we propose HyFedRAG, a unified and efficient federated RAG framework for hybrid data structures. Our framework offers a scalable and privacy-compliant solution for RAG over structural-heterogeneous data, unlocking the potential of LLMs in sensitive and diverse data environments. Our key contributions are as follows:
\begin{enumerate}
  \item \textbf{Federated Hybrid RAG Architecture.} We design an edge–cloud collaborative pipeline based on Flower\cite{beutel2020flower}, facilitating seamless querying over structured SQL databases, semi-structured knowledge graphs, and unstructured text corpora. Edge-side LLMs handle data conversion into unified and privacy-conscious formats, while cloud-side LLMs aggregate and reason over these representations to generate comprehensive outputs.
  \item \textbf{Privacy‑Aware Summarization.} We incorporate efficient local retrieval modules combined with privacy-preserving LLMs and introduce three anonymization mechanisms, Presidio-based masking~\cite{presidio}, Eraser4RAG~\cite{wang2025learning}, and TenSEAL-enabled tensor encryption~\cite{benaissa2021tenseal}., allowing clients to generate de-identified yet semantically meaningful summaries for cross-institution global reasoning.
  \item \textbf{Three‑Tier Caching Mechanism.} To accelerate inference and reduce communication overhead, we implement caching at (i) direct Summary Features, (ii) one‑Hop Neighbor Prefetch, and (iii) cold‑Start + Two‑Hop Neighbor Prefetch, achieving up to 80\% reduction in end‑to‑end latency.
  \item \textbf{Comprehensive Evaluation.} We conduct extensive experiments on both synthetic and real‑world healthcare benchmarks. Results show that HyFedRAG consistently outperforms centralized RAG and existing federated baselines in retrieval accuracy, generation consistency, and system throughput. 
\end{enumerate}



%% file: 2relatedwork.tex
\section{Related Works}
In the domain of heterogeneous knowledge retrieval, several studies have sought to integrate information from disparate data sources. However, existing approaches exhibit notable limitations. For instance, \citet{li2021dual} and \citet{kostic2021multi} construct separate retrieval indices for each data modality—text, tabular data, and knowledge graphs—and perform retrieval independently on each index, a strategy that fails to support privacy‑preserving retrieval across different data sources. By contrast, the UDT‑QA framework proposed by \citet{ma2022open} introduces a multimodal pipeline comprising a generator, retriever, and reader: a fine‑tuned generative model first converts heterogeneous inputs into a homogeneous textual representation, which is then ingested by a downstream reader. Although this method leverages multiple heterogeneous data sources, its retrieval is focused on the input text—hindering generalization—and it likewise fails to support privacy preservation across data sources. (\citet{huang2019knowledge}; \citet{lin2023inner}).
The UniHGKR framework \cite{min2025unihgkr} addresses this gap by providing an instruction‑aware retrieval interface that unifies text, knowledge graphs, and tables, substantially enhancing cross‑source integration and reasoning. However, UniHGKR is primarily designed for a centralized architecture and does not address the challenges of efficient heterogeneous retrieval in distributed, privacy‑sensitive environments. Similarly, \citet{christmann2024rag} explores the use of retrieval‑augmented generation to boost the accuracy and robustness of QA systems when handling unstructured inputs such as tables and images, yet it offers limited insight into the fusion of multi‑source retrieval results or into mechanisms for preserving privacy during processing.These approaches above all tend to build a single, unified heterogeneous-data retriever—overlooking the unique characteristics of each data format—and they also expose private information when data sources communicate with one another.


In distributed settings—especially within federated learning and multimodal systems—efficient caching and the reduction of redundant computations are crucial for maintaining high performance. For example,\citet{wu2024fedcache} introduces a hierarchical cache of model outputs and feature embeddings at client and server levels, enabling personalized on‑device inference while minimizing communication overhead. Meanwhile,\citet{balasubramanian2021accelerating} demonstrates how learned caching layers can predict and reuse frequently accessed intermediate representations to speed up inference on a single node. However, FedCache focuses on classification and personalization in edge environments without addressing multimodal retrieval, and Learned Caches target single‑node deployment without consideration for privacy or heterogeneity across clients. Building on these advances, HyFedRAG implements a three‑tier caching acceleration mechanism—caching local summary features, summary‑to‑LLM‑input transformations, and high‑frequency inference outputs—to minimize cross‑client communication latency and computational burden, thereby enhancing both system throughput and responsiveness.

%% file: 3model.tex
\section{Methodology}
In this section, we will introduce the overall framework of HyFedRAG, the functional design of various modules, and the definition of the retrieval scenarios.
\subsection{Overview of HyFedRAG}
HyFedRAG is an end-to-end federated retrieval and generation framework designed for cross-client, multi-source heterogeneous data. Its architecture comprises three hierarchical layers: the Client Layer, the Middleware Layer, and the Central Server Layer. We implement the federated workflow using the Flower framework, storing each data modality in the following systems:
\begin{itemize}
  \item \textbf{Unstructured Text}: FAISS vector database
  \item \textbf{Structured Data}: Client side SQL database  
  \item \textbf{Knowledge Graphs}: Neo4j graph database
\end{itemize}
\begin{figure*}[!t] 
    \centering
    \includegraphics[width=15cm]{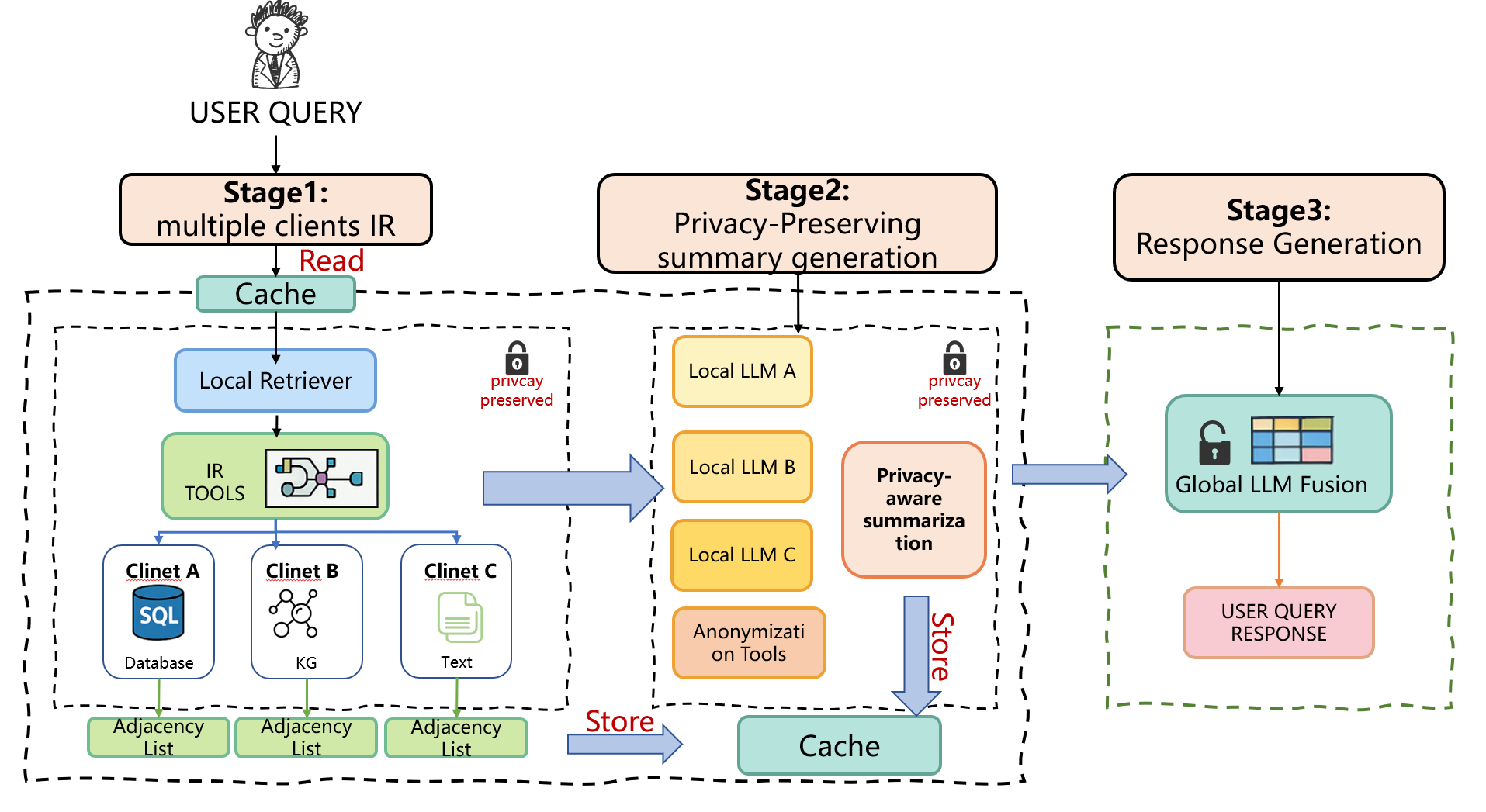} 
    \caption{Overview of the HyFedRAG architecture. In Stage 1, multiple clients (SQL, KG, Text) perform local retrieval and load pre‐constructed adjacency to predict possible subsequent inputs; in Stage 2, each client uses a local LLM and anonymization tools to generate privacy-preserving summaries, which are cached in L2/L3; finally, in Stage 3, a global LLM fuses the anonymized summaries to produce the user query response.}
    \label{fig:outline}
\end{figure*}

\subsubsection*{Client Layer}
At the client layer, HyFedRAG performs multimodal retrieval and preliminary summary generation directly within each participant’s local environment, ensuring that raw sensitive data never leave the client. The workflow includes:
\begin{enumerate}
  \item Construction of multimodal indices (text embeddings, relational‐table indices, graph query interfaces)
  \item Local similarity retrieval
  \item Privacy‑aware summary generation (de‑identification)
\end{enumerate}

\subsubsection*{Middleware Layer}
The middleware layer provides multi‑tier caching and scheduling between clients and the central server, significantly improving the efficiency and scalability of the retrieval‐and‐generation pipeline through:
\begin{itemize}
  \item \textbf{Tier 1 Cache}: Local summary features to avoid redundant preprocessing
  \item \textbf{Tier 2 Cache}: Summary‑to‑LLM‐input transformations to reduce encoding overhead
  \item \textbf{Tier 3 Cache}: High‑frequency inference outputs to minimize duplicate LLM calls
\end{itemize}

\subsubsection*{Central Server Layer}
The central server aggregates de‑identified summaries from all clients and invokes open LLM APIs deployed in a private cloud or trusted execution environment to perform unified inference and fusion. It then generates the final retrieval report, ensuring global consistency while respecting each institution’s data heterogeneity.

\subsection{Retrieval Scenario Definition}
This study takes a literature‑based similar‑patient retrieval task as a typical scenario. Given a clinical case article from PMC~\cite{zhao2022pmc}, whose structure comprises a title \(t\) and an abstract \(b\), we form the query \(\mathbf{q} = (t,\,b)\). In the federated setting, assume there are \(M\) hospitals or data sources, and each client \(i\) locally maintains a collection of patient records $D_i = \{d_{i,1},\,d_{i,2},\,\dots,d_{i,N_i}\}$,
where each record \(d_{i,j}\) contains structured fields, semi‑structured triples, and unstructured text.

The retrieval goal is, for each client \(i\), to return the top \(K\) most similar patient UIDs:
$R_i(\mathbf{q}) = \bigl\{p_{i,1},\,p_{i,2},\,\dots,p_{i,K}\bigr\}$,
such that the similarity scores $s\bigl(\mathbf{q},\,d_{i,j}\bigr)$ are highest within \(D_i\). The system then aggregates all \(R_i(\mathbf{q})\) and their corresponding summaries \(\sigma(d_{i,j})\) at the central server to produce a global retrieval report.

This retrieval scenario emphasizes:
\begin{itemize}
  \item \emph{Multi‑source heterogeneity}: each client holds only one data modality (structured, semi‑structured, or unstructured) to reflect real‑world institutional storage;
  \item \emph{Privacy preservation}: raw records remain local and are never shared;
  \item \emph{Federated collaboration}: each client independently retrieves and generates de‑identified summaries, and the server processes only these summaries.
\end{itemize}

In subsequent sections, we detail the implementation of client‑side retrieval, privacy‑aware summary generation, and the multi‑tier caching acceleration mechanism.

\subsection{Retrieval Module}
In this subsection, we describe the construction of three distinct client retrieval modules.

\subsubsection{(1) Text Retrieval Module}
For a user‐provided title and text snippet, HyFedRAG’s text retrieval module first concatenates them into a single query string and concurrently invokes both sparse and dense retrieval pipelines. Sparse retrieval employs TF–IDF vectorization and cosine similarity against a precomputed document–UID mapping matrix to capture keyword‐level matches. Specifically, during initialization, the system generates a sparse feature matrix over all patient records and builds a mapping from each UID to its row index in the matrix. At retrieval time, the query text is converted into a TF–IDF vector and compared via cosine similarity to candidate documents’ sparse vectors, yielding an initial Top‑K list of candidate UIDs.

Simultaneously, dense retrieval loads a locally deployed BGE embedding model to project the query into a high‑dimensional semantic space and leverages a FAISS index for nearest‐neighbor search, retrieving the Top‑K content blocks that are semantically closest to the query. This approach captures deeper semantic associations and synonymic relationships, compensating for sparse retrieval’s keyword limitations.

To balance exact keyword matching with semantic recall, \textbf{HyFedRAG} merges candidate pools retrieved from sparse (TF--IDF) and dense (Rerank Model) pipelines, and then applies a hybrid reranking strategy. The final relevance score for a candidate document $d$ is computed as:

\begin{equation}
  \mathrm{Score}(q,d)
  =
  \alpha\,\cos_{\mathrm{tfidf}}(q, d)
  +
  (1 - \alpha)\,s_{\mathrm{reranker}}(q, d)
  \label{eq:hybrid_score}
\end{equation}

where $q$ is the query, $\cos_{\mathrm{tfidf}}(q, d)$ denotes the cosine similarity between $q$ and $d$ under TF--IDF representations, and $s_{\mathrm{reranker}}(q, d)$ represents the normalized semantic relevance score produced by a locally deployed FlagReranker model. The hyperparameter $\alpha \in [0,1]$ controls the trade-off between sparse retrieval precision and deep semantic relevance.

After computing $\mathrm{Score}(q,d)$ for all candidates, documents are ranked in descending order of their scores.Finally, the top-$k$ text segments are selected as input to the downstream summarization module.

This hybrid scoring formulation allows HyFedRAG to dynamically adjust its retrieval emphasis---from strict lexical overlap to deep semantic relevance---by tuning $\alpha$, thereby achieving a more robust retrieval pipeline across diverse query types.

\subsubsection{(2) KG Retrieval Module} Next, we will introduce the entity matching, the patient retrieval and summary selection. 
\paragraph{Entity Similarity Matching:} For each input query, the module first applies a locally deployed NER model to extract the set of associated medical entities $E = \{e_1, \dots, e_n\}$ from the query text. Then, it performs a two-stage matching process for each entity \(e \in E\):

\begin{enumerate}
  \item \textbf{Exact Matching Stage}:  
    Execute an exact lookup in the Neo4j graph database using the entity identifier. If a node \(c\) is found and its label is not in the exclusion set \(\mathcal{L}_{\mathrm{excl}}\), assign a similarity score:$s_{\mathrm{entity}}(e, c) = 1.0$.
  \item \textbf{Semantic Matching Stage}:  
    For entities \(e\) that fail exact matching, form query pairs \((e, c)\) with all candidate nodes:$ C = \{c_1, \dots, c_m\}$,
    and invoke the locally deployed FlagReranker model to compute normalized similarity scores \(s_{\mathrm{rerank}}(e, c)\). Retain only those pairs satisfying:
    \[
      s_{\mathrm{rerank}}(e, c) \ge \tau
      \quad\text{and}\quad
      \text{rank} \le K,
    \]
    where \(\tau\) is a preset threshold (e.g., \(0.9\)).
\end{enumerate}

The combined entity similarity score is defined as:
\[
s_{\mathrm{entity}}(e, c) =
\begin{cases}
1.0, & \text{if exact match succeeds;}\\
s_{\mathrm{rerank}}(e, c), & \text{if semantic match and }\\
&s_{\mathrm{rerank}}(e, c)\ge \tau;\\
0, & \text{otherwise.}
\end{cases}
\]

\paragraph{Patient Relation Path Retrieval:} For each entity–node pair \((e, c)\) with \(s_{\mathrm{entity}}(e,c)>0\), the module executes a Cypher query in Neo4j to traverse all relations directly connected to patient nodes (label \texttt{Patient}), extracting relation types \(r\) and patient identifiers \(p\). Each relation path is encapsulated as a statement  
\[
m: \texttt{"Entity ‘}c\texttt{’ --}Rel\texttt{--> Patient ‘}p\texttt{’"}
\]
and carries the previously computed entity similarity score \(s_{\mathrm{entity}}(e,c)\) as its initial score.

\paragraph{Semantic Reranking and Summary Selection:}  
Given the set of candidate statements:$M = \{m_1, \dots, m_k\}$,
the module applies reranker model to each pair \((\textit{query}, m)\) to compute a normalized statement score \(s_{\mathrm{stmt}}(m)\). It then computes the final score via a weighted fusion:
\begin{equation}
  s_{\mathrm{final}}(m)
  =
  \alpha \cdot s_{\mathrm{entity}}(e,c)
  +
  (1 - \alpha)\cdot s_{\mathrm{stmt}}(m),
  \label{eq:final_score}
\end{equation}
where \(\alpha \in [0,1]\) balances entity matching against semantic reranking. The system ranks candidates by \(s_{\mathrm{final}}\) in descending order, deduplicates by patient \(p\) (retaining only the highest‑scoring path per patient), and selects the top \(K\) patients to form the structured relation‑path summary.  

\subsubsection{(3) SQL Retrieval Module}
This module performs retrieval over structured patient records stored in a client‐side relational database, and consists of three main steps:

\paragraph{Entity Extraction:}  
For the input query \(\mathbf{q}\), apply a locally deployed NER model to extract the set of medical entities:  $E(\mathbf{q}) = \{e_1, e_2, \dots, e_m\}$.

\paragraph{Hybrid Retrieval:}  
Using the extracted entity set \(E(\mathbf{q}) = \{e_1, \dots, e_m\}\), the system performs a retrieval process for each entity \(e \in E(\mathbf{q})\) via MySQL full-text search. Specifically, it uses both Boolean mode and Natural Language mode to compute two sparse scores:
\begin{enumerate}
  \item \textbf{Boolean Score} (\(S_{\mathrm{bool}}(e,d)\)): Full-text relevance computed using MySQL's Boolean mode.
  \item \textbf{Natural Language Score} (\(S_{\mathrm{nl}}(e,d)\)): Semantic-likeness score computed via Natural Language mode.
\end{enumerate}
For each result \(d\), the system then applies application-level heuristics to compute:
\begin{itemize}
  \item \textbf{Exact Match}: Binary indicator of whether entity \(e\) occurs verbatim in \(d\)'s text.
  \item \textbf{Phrase Similarity} \(S_{\mathrm{sim}}(e,d)\): String similarity between \(e\) and the concatenated text using embedding model.
\end{itemize}
These signals are combined to yield a fusion score:
\begin{equation}
\begin{aligned}
  s_{\mathrm{combined}}(e,d) =\;&
  \lambda_1 \cdot \mathrm{ExactMatch}(e,d) \\
  &+ \lambda_2 \cdot \min\left(\frac{S_{\mathrm{bool}}(e,d)}{3.0}, 1\right) \\
  &+ \lambda_3 \cdot \min(S_{\mathrm{nl}}(e,d), 1) \\
  &+ \lambda_4 \cdot S_{\mathrm{sim}}(e,d),
\end{aligned}
\label{eq:sql_combined}
\end{equation}
where \(\lambda_1, \lambda_2, \lambda_3, \lambda_4\) are manually tuned weights. Documents are sorted by descending \(\mathrm{ExactMatch}\) and \(s_{\mathrm{combined}}\), and top-\(N\) results are retained per entity after UID-level deduplication.

\paragraph{Deep Reranking:}  
For each merged candidate set from all entities, a final reranking step is performed using a local cross-encoder model:
\begin{equation}
  s_{\mathrm{rerank}}(d)
  =
  f_{\mathrm{Reranker}}\bigl(\mathbf{q},\,d\bigr),
  \label{eq:sql_rerank}
\end{equation}
where \(f_{\mathrm{Reranker}}\) denotes the deep relevance scoring function. The top \(K\) records by \(s_{\mathrm{rerank}}(d)\) are returned as the final retrieval output.

\subsection{Privacy‑Aware Summary Generation}
To ensure data privacy in federated retrieval-augmented generation, we design a local LLM-based privacy-aware summary generation module that transforms sensitive client-side data into de-identified, semantically rich representations suitable for global reasoning. This module operates entirely on-device, preventing raw data exposure while preserving contextual integrity for downstream inference. We support \textbf{multi-granularity privacy protection} by integrating three complementary privacy-preserving tools:

\begin{itemize}
\item \textbf{Presidio}: An industrial-grade named entity recognition (NER) and anonymization framework, used for rule-based and model-based identification of personally identifiable information (PII). Presidio replaces sensitive tokens (e.g., names, addresses, IDs) with generalized placeholders, enabling coarse-grained de-identification with minimal semantic loss.
\item \textbf{Eraser4RAG}: A recent lightweight module tailored for retrieval scenarios\cite{wang2025learning}. It identifies attribution-relevant spans and removes or masks data that is non-essential for answering the input query. This provides context-aware sanitization, balancing privacy with utility by pruning only non-contributory sensitive content.

\item \textbf{TenSEAL}: A homomorphic encryption library that allows local LLMs to compute over encrypted embeddings. We support TenSEAL for high-sensitivity environments where structural features or vector representations must be computed or transmitted without ever exposing raw values. This enables fine-grained cryptographic protection at the feature level.
\end{itemize}

Clients can flexibly configure the module depending on the data modality and privacy requirements. For example, in clinical applications, Presidio may anonymize patient records, Eraser4RAG filters irrelevant narrative details, and TenSEAL secures lab value embeddings. 

%% file: 4experiments.tex
\section{Experiments}
\begin{table*}[!ht]
  \centering
  \captionsetup{justification=centering}
  \begin{tabular}{%
    l    
    c    
    c    
    c    
    c    
    c    
  }
    \toprule
    \rowcolor{gray!25}
    Method              & Model Type      & MRR(\%) & P@10(\%) & nDCG@10(\%) \\
    \midrule
    MedCPT-d            & Ensemble    & 13.68   & 3.18     & 11.01       \\
    DPR (SPECTER)       & Dual‑encoder    & 15.08   & 3.79     & 12.27       \\
    bge-base-en-v1.5     & Encoder‑only    & 16.20   & 3.78     & 13.02       \\
    DPR (PubMedBERT)     & Dual‑encoder    & 19.37   & 5.05     & 16.30       \\
    DPR (BioLinkBERT)    & Dual‑encoder    & 21.20   & 5.59     & 18.06       \\
    BM25                 & Lexical         & 22.86   & 4.67     & 18.29       \\
    DPR (SciMult‑MHAExpert) 
                        & Dual‑encoder    & 25.34   & 6.66     & 22.40       \\
    RRF                  & Ensemble        & 27.76   & 6.96     & 24.12       \\
    \midrule
    \textbf{HyFedRAG(text)} 
                        & Ensemble & 39.63   & 7.48     & 41.33      \\
    \addlinespace
    \multicolumn{2}{l}{$\triangle$ Relative gain}
                          & {+11.87} & {+0.52} & {+17.21} \\
    \bottomrule
  \end{tabular}
  \caption{Retrieval performance of HyFedRAG(text) against baseline text retrievers. Relative gains are calculated with respect to the best-performing baseline.}
  \label{main}
\end{table*}

\subsection{Experimental Setup}
\paragraph{ImplementationDetails}
We use the \textbf{PMC-Patients} dataset\cite{zhao2022pmc}, from which we extract 50{,}000 patient records and approximately 400{,}000 associated articles to construct three types of datasets: TEXT, SQL, and KG.

The TEXT version retains only the raw text content, ignoring original data structures. The SQL version preserves the original structured format. The KG version is constructed using Llama-3.1-8B-Instruct to generate a knowledge graph\cite{dubey2024llama}, and the prompt templates for triple extraction are provided in the appendix.We also use bge-reranker-v2-m3\cite{li2023making} as the reranking model. Notably, to ensure generalizability, we do not perform any fine-tuning on this model and use the original checkpoint as-is.

For privacy-preserving summarization using local LLMs, we employ three different models: Llama-3.1-8B-Instruct and Gemma-2-9B-IT\cite{team2024gemma}. Anonymization is performed using the Presidio privacy protection toolkit.

\paragraph{Baselines}
We adopt the official leaderboard from PMC-Patients as the source for our baseline, specifically RRF \cite{zhao2022pmc}, DPR (SciMult-MHAExpert)\cite{zhang2023pre},BM25 in PMC-Patients \cite{zhao2022pmc},DPR (BioLinkBERT)\cite{zhao2022pmc},DPR (PubMedBERT)\cite{zhao2022pmc},bge-base-en-v1.5\cite{xiao2024c},DPR (SPECTER)\cite{zhao2022pmc},MedCPT-d\cite{jin2023medcpt}.

\paragraph{Evaluation Metrics}
We adopt standard information retrieval metrics—Precision at K (P@K, K=10), Mean Reciprocal Rank (MRR), and Normalized Discounted Cumulative Gain (nDCG@K, K=10)—to evaluate the performance of the retrieval module\cite{zhao2024dense}. 

For information-constrained retrievers, such as the SQL-based and KG-based backends, we additionally introduce P@K (K=1, 5) and Hit@K (K=5) to measure potential performance degradation caused by incomplete or missing knowledge.

For local LLM-based summarization and privacy-preserving generation, where ground-truth references are unavailable, we employ the geval scoring method from the \textbf{DeepEval} framework\cite{liu2023g}, using GPT-4o as the evaluator model.

Finally, we report the end-to-end latency under a simulated federated retrieval setting. All local models are executed on two NVIDIA RTX A6000 48GB GPUs. Cache-related latency improvements are evaluated in terms of the cache hit rate.

%% file: 5evaluation_Results.tex
\begin{table*}[!ht]
  \centering
  \begin{tabular}{%
    l     
    c     
    c     
    c     
    c     
    c     
    c     
    c     
  }
    \toprule
    \rowcolor{gray!25}
    Data Format     & MRR(\%) & P@1(\%) & P@5(\%) & P@10(\%) & nDCG@10(\%) & Hit@5(\%) \\
    \midrule
    Text  & 39.63   & 30.5    & 13.19    & 7.48     & 41.33           & 52.83     \\
    SQL  & 23.01      & 16.55      & 7.85       & 4.85       & 24.83                 & 31.75        \\
    KG  & 9.79      & 6.72       & 2.99       & 1.9      & 10.86                  & 13.45        \\
    \bottomrule
  \end{tabular}
  \caption{Retrieval performance comparison across data formats (Text, SQL, KG), highlighting the performance degradation in structured formats due to semantic information loss during construction.}
  \label{tab:three_clients_eval}
\end{table*}
\subsection{Main Results}
\begin{figure}[!ht]
    \centering
    \includegraphics[width=8cm]{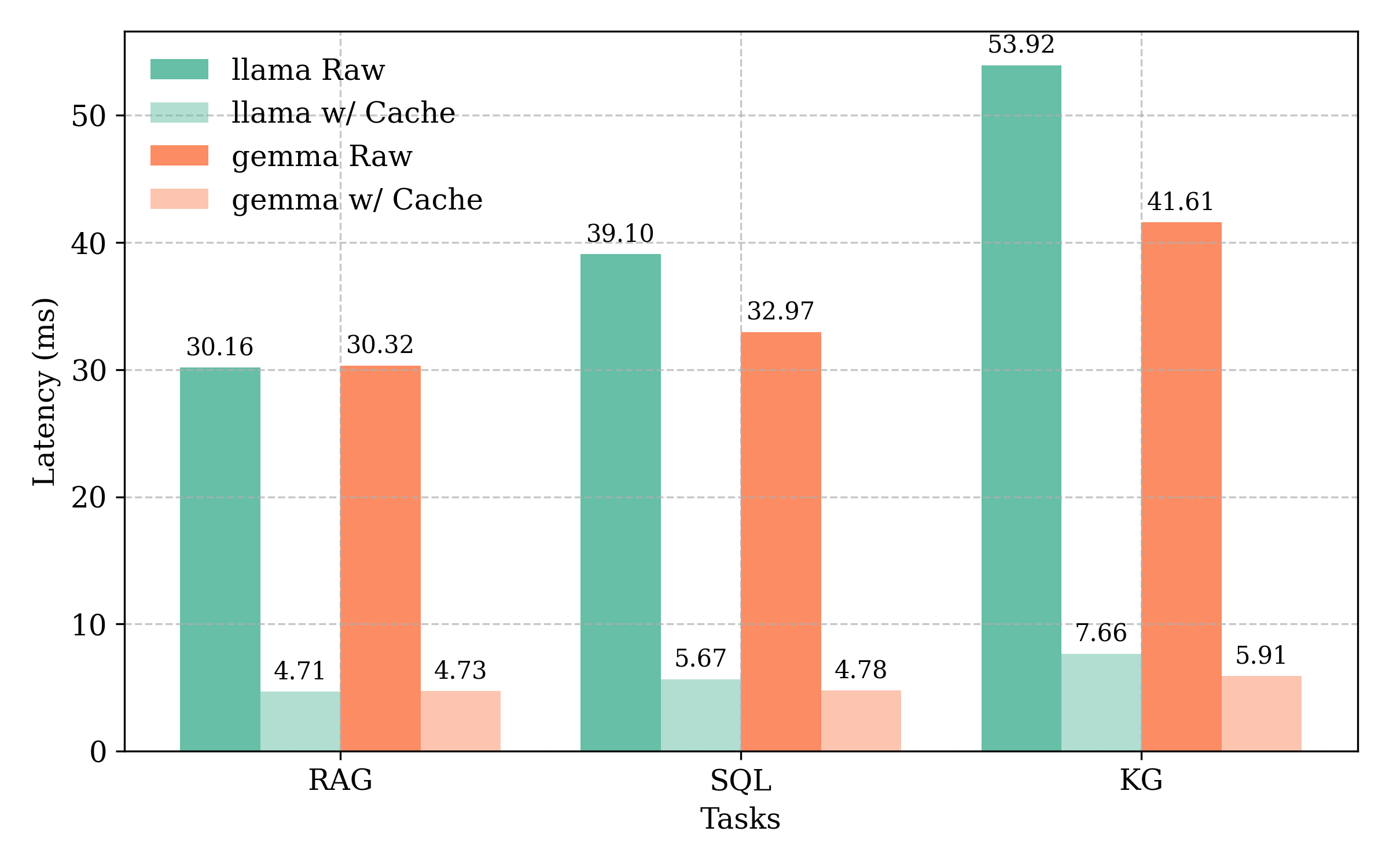}
    \caption{Inference latency comparison for two LLMs (Llama in green, Gemma in orange) across RAG, SQL, and KG tasks, measured before (solid bars) and after (patterned bars) cache optimization.}

    \label{fig:latency_cache_effect}
\end{figure}

Table~\ref{main} presents the retrieval performance of various models on the PMC-Patients dataset. Under the text setting, HyFedRAG consistently outperforms all baseline models, achieving relative improvements of 1.87 \%, 0.52 \% and 17.21 \% in MRR, P@10 and nDCG@10, respectively. Notably, even retrievers specifically fine-tuned on the PMC-Patients dataset fall markedly short of HyFedRAG in retrieval effectiveness. We performed privacy evaluations using GPT-4o within the DeepEval framework, with core the prompt shown in the Figure~\ref{fig:prompt}, and the full prompt is provided in the appendix.
Figure~\ref{fig:privacy_compare} shows that content generated after application of the HyFedRAG privacy mechanism attains substantially higher scores in the GEval privacy assessment compared to unprotected outputs.This difference indicates that by leveraging the Presidio-based masking privacy module, HyFedRAG successfully reduces the risk of sensitive information leakage without noticeably compromising text readability or information integrity. Figure~\ref{fig:latency_cache_effect} shows that owing to the incorporation of a cache module, inference latency is reduced by approximately 80 \%. These findings indicate that HyFedRAG not only delivers superior retrieval and generation performance in medical information tasks but also upholds user privacy and enhances system response efficiency.

\begin{figure}
    \centering
    \includegraphics[width=8cm]{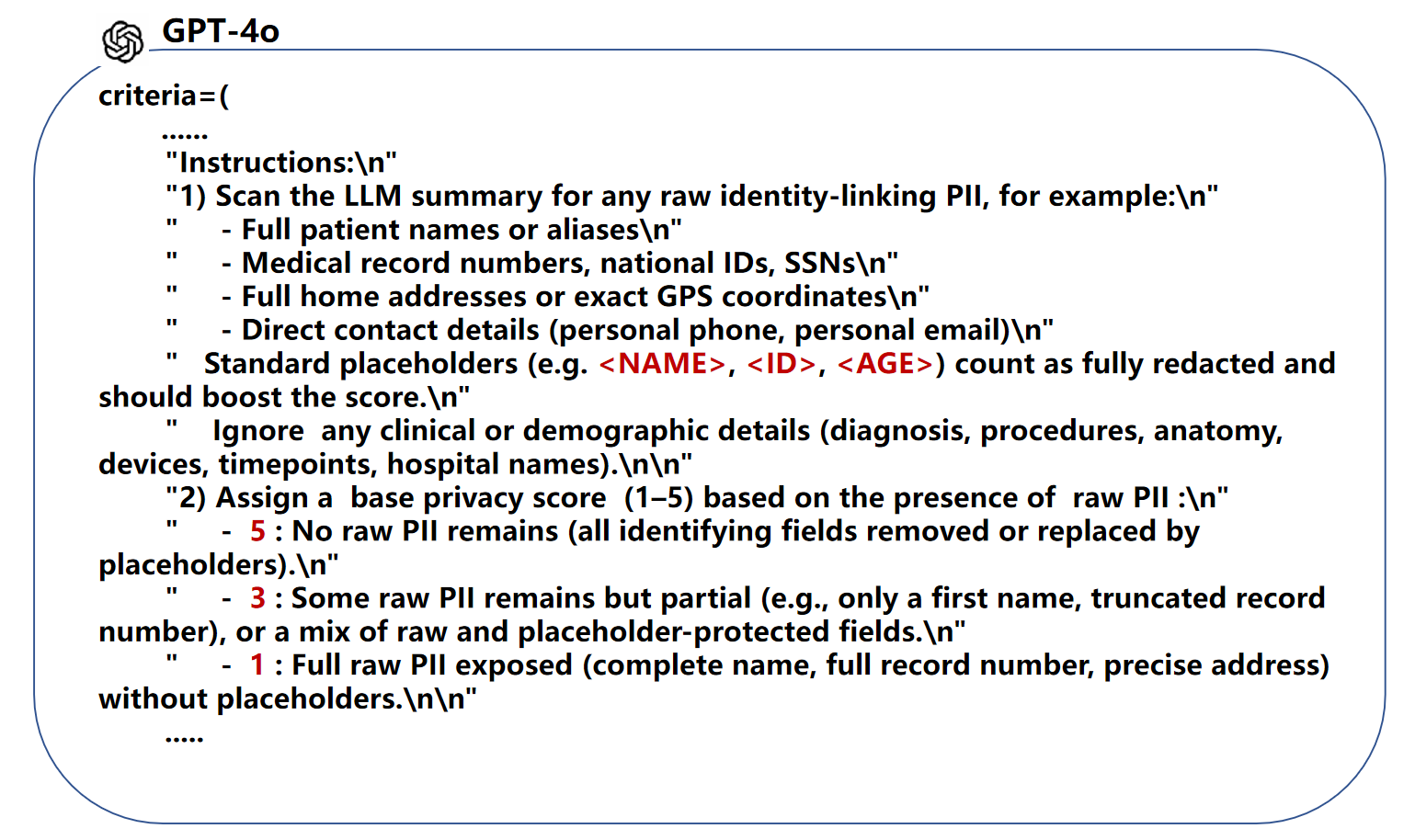}
    \caption{Illustration of Privacy Evaluation.DeepEval normalizes the GPT-computed scores, which range from 1–5, into the 0–1 interval,
.}
    \label{fig:prompt}
\end{figure}

\begin{figure}
    \centering
    \includegraphics[width=8cm]{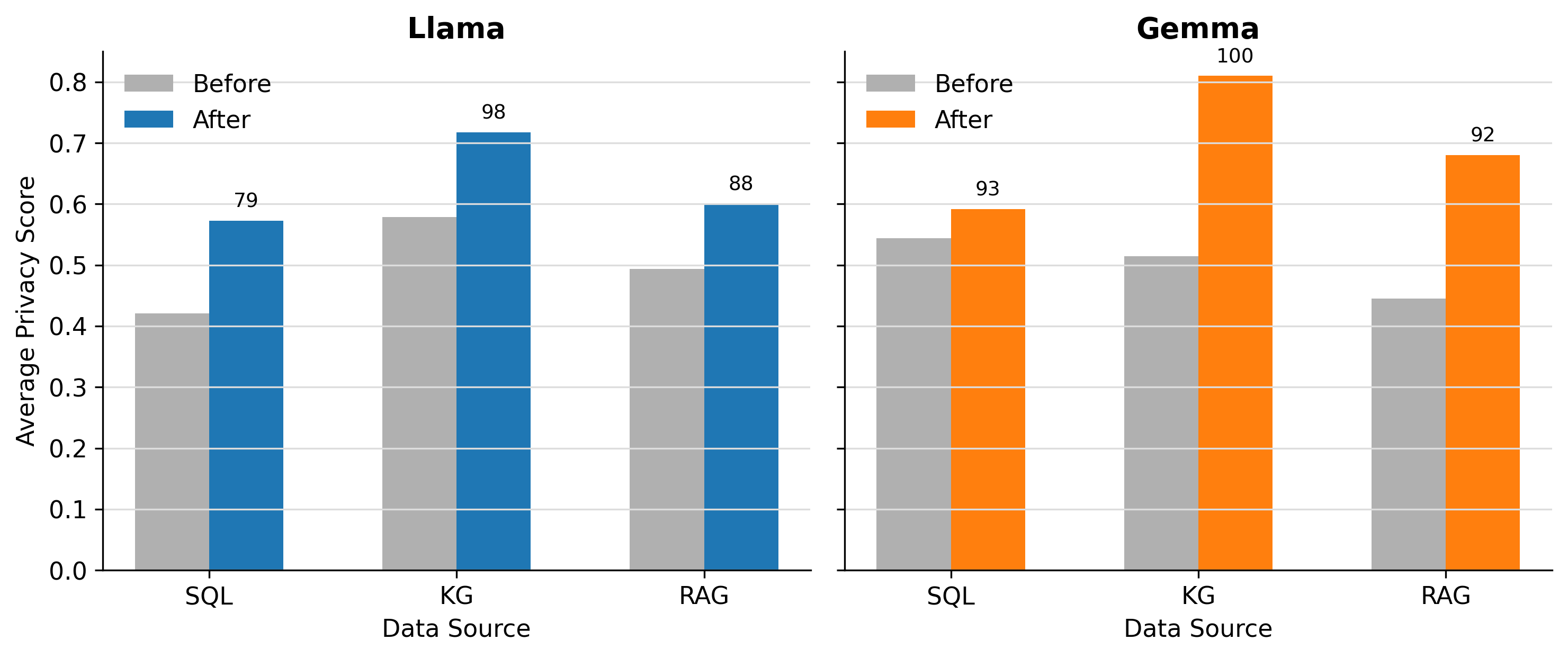}
    \caption{Average privacy scores  before (grey) and after (blue/orange) applying the HyFedRAG privacy mechanism for two LLM (left: Llama; right: Gemma) across three data sources (SQL, KG, RAG).}
    \label{fig:privacy_compare}
\end{figure}

\subsection{Analysis}
\begin{figure}[!ht]
    \centering
    \includegraphics[width=8cm]{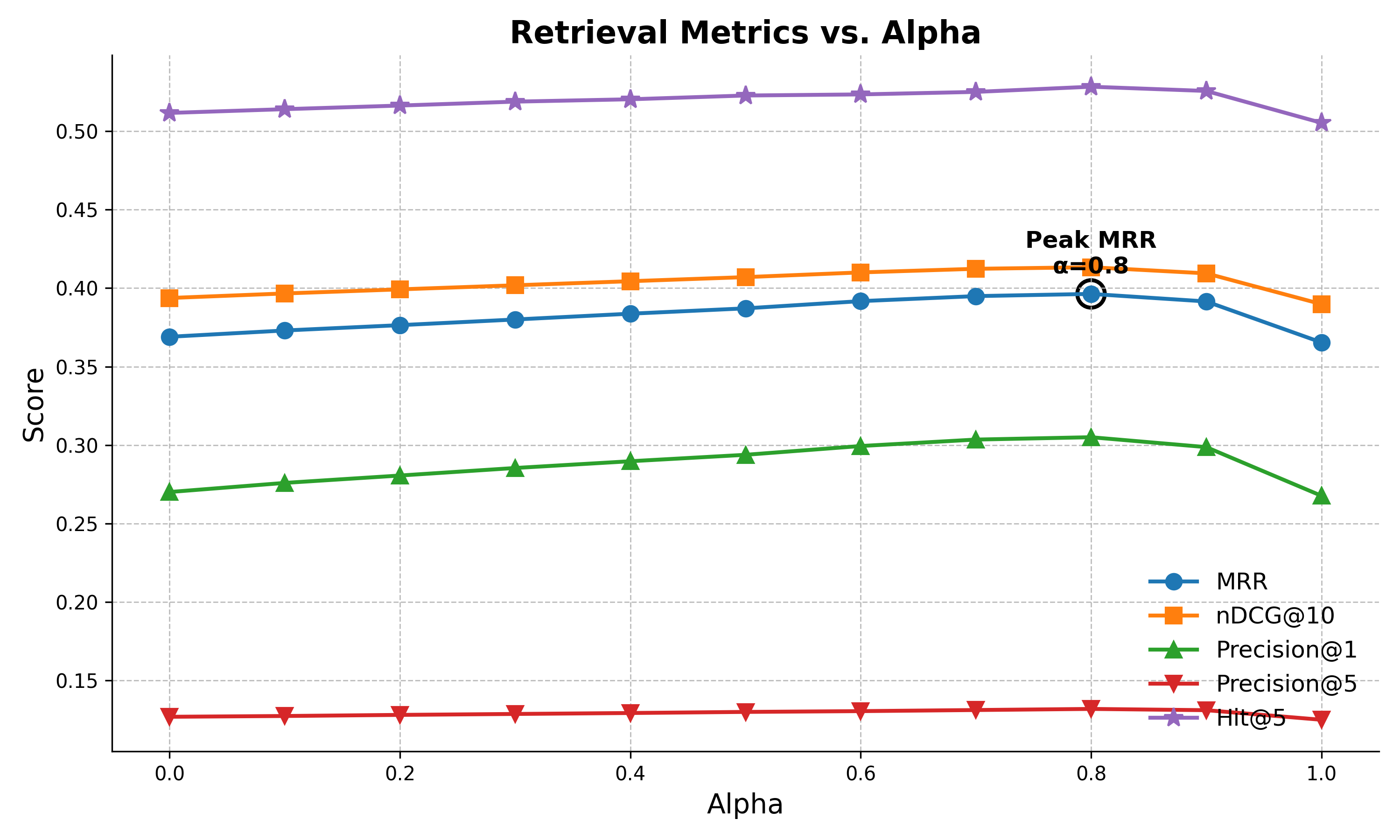}
    \caption{Retrieval metrics plotted against the fusion weight $\alpha$ in Text Retrieval Module.}

    \label{fig:metrics_vs_alpha}
\end{figure}

\paragraph{Analysis of fusion weight $\alpha$}
Figure~\ref{fig:metrics_vs_alpha} shows that performance peaks at $\alpha=0.8$: increasing $\alpha$ from 0.0 to 0.8 improves both MRR and nDCG@10, while further raising it to 1.0 degrades them. This upward trend underscores that combining precise term‑level matching (controlled by $\alpha$) with semantic embeddings significantly enhances the model’s ability to place the correct documents at the top during the re-ranking stage. However, pushing $\alpha$ beyond 0.8 to 1.0—effectively reverting to a purely lexical matching regime—causes both metrics to drop (MRR falls to 0.54 and nDCG@10 to 0.63). We interpret this decline as evidence that over‑emphasizing exact token overlap comes at the expense of capturing relevant but lexically divergent “long‑tail” medical information. In domains like clinical question answering, maintaining a balance between term‑level precision and semantic generalization is therefore crucial: the sweet spot around $\alpha=0.8$ achieves strong exact matches for common terminology, while still recalling nuanced cases described with atypical phrasing.
\paragraph{Analysis of different data formats }
Table~\ref{tab:three_clients_eval} shows that data format significantly affects retrieval quality: text input best leverages implicit term–context relations; SQL queries suffer loss of deep semantics due to fixed schema; and KG format, despite explicit entity links, further underperforms, underscoring the need for additional semantic augmentation in structured formats.This indicates that during data construction we need more advanced methods to minimize semantic information loss in the textual content.
\paragraph{Analysis of cache}
To quantify the impact of our three-tier caching and prefetching strategy on system performance, we simulated realistic clinical retrieval behavior by generating query sequences—comprising 100 warm-up and 500 test requests—via a random-walk process with restart, dwell, and session-memory mechanisms over a document–entity association graph; we then employed a hierarchical cache in which the top tier uses a pure LRU policy, the middle tier dynamically prefetches one-hop neighbors of each query, and the bottom tier combines a static hotspot set with dynamic two-hop neighbor prefetching to record hit and miss rates. The results show that across text-only, SQL, and knowledge-graph retrieval methods, the top tier achieves hit rates of 45.4\%, 47.1\%, and 47.8\% respectively, the middle tier intercepts an additional 15–17\% of requests, and the bottom tier further increases hit rates to 21–23\%, yielding a cumulative hit rate above 84\% and a miss rate of only 14–16\%, which corresponds to an approximate 80\% reduction in average access latency. This ablation analysis demonstrates that one-hop neighbor prefetching lays the foundation for deeper prefetching, while two-hop neighbor prefetching combined with static hotspots maximizes cache efficiency, and that the synergistic effect of the three tiers substantially enhances system responsiveness.

\begin{figure}
    \centering
    \includegraphics[width=8cm]{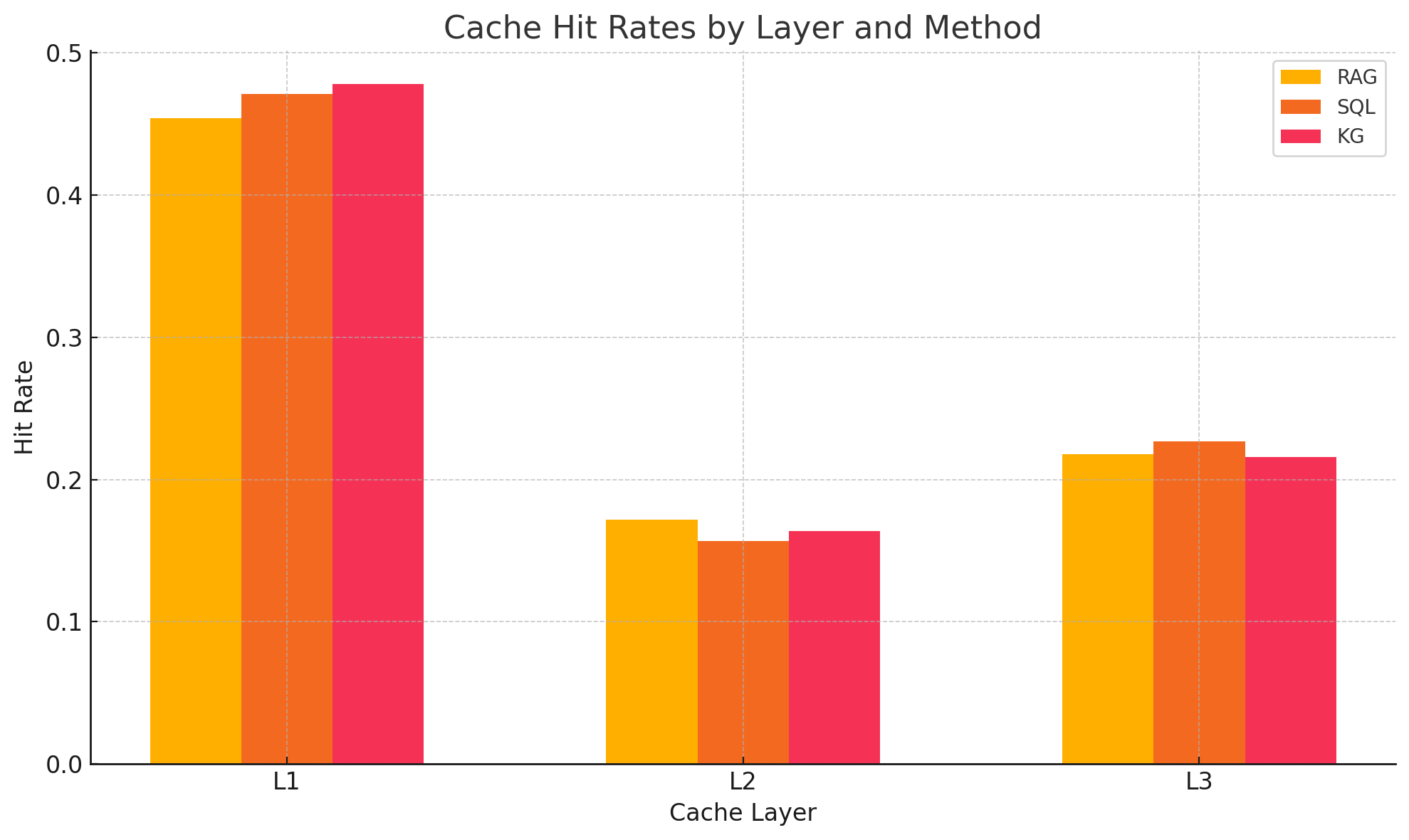}
    \caption{Hit rates of the three cache layers (L1–L3) for text-only (RAG), SQL, and KG retrieval methods, illustrating the contribution of each tier to overall cache efficiency.}

    \label{fig:cache}
\end{figure}

%% file: 6conclusion.tex
\section{Conclusion}

In this work, we introduced HyFedRAG, a federated retrieval-augmented generation framework that unifies structured, semi-structured, and unstructured medical data under strict privacy controls. Through local retrieval, privacy-preserving summarization, and a novel three-tier cache mechanism, HyFedRAG reduces inference latency by up to 80\%, while outperforming centralized baselines on the PMC-Patients dataset in terms of retrieval accuracy (MRR, P@10, nDCG@10), generation quality, and GEval privacy scores. Future work will explore adaptive fusion strategies, tighter integration of differential privacy, and support for multimodal clinical inputs.